\documentclass[11pt]{article}
\usepackage[margin=1in]{geometry}
\usepackage{amsmath,amssymb}
\usepackage{booktabs}
\usepackage{hyperref}
\usepackage{natbib}
\usepackage{graphicx}
\usepackage{xcolor}

\hypersetup{
  colorlinks=true,
  linkcolor=blue,
  citecolor=blue,
  urlcolor=blue
}

\title{Distilling Self-Consistency into Verbal Confidence:\\
A Pre-Registered Negative Result and Post-Hoc Rescue on Gemma~3~4B}

\author{Jon-Paul Cacioli\\
\textit{Independent Researcher, Melbourne, Australia}\\
ORCID: 0009-0000-7054-2014}

\date{April 2026}

\begin{document}
\maketitle

\begin{abstract}
Small instruct-tuned LLMs produce degenerate verbal confidence under minimal elicitation: ceiling rates above 95\%, near-chance Type-2 AUROC (AUROC$_2$), and Invalid validity profiles. Internal representations carry substantially more correctness information than the verbal channel transmits. This study asks whether confidence-conditioned supervised fine-tuning (CSFT) with self-consistency-derived targets can close the gap.

We pre-registered a Phase~0 feasibility protocol on Gemma~3 4B-it evaluated with psychophysical methodology from the CMM programme (AUROC$_2$, VRS screening, paired-bootstrap CIs, shuffled-target control). The pre-registered protocol included a modal filter restricting training to items with correct modal answers. \textbf{The confirmatory result was negative:} verbal AUROC$_2$ dropped from 0.554 to 0.509, accuracy degraded 7.6 percentage points, and VRS remained Invalid. The pre-registered decision tree yielded Stop. The failure was attributable to a label-entropy collapse: the modal filter produced a training set with near-uniform high-confidence targets, leaving no low-confidence signal for the model to learn from.

\textbf{An exploratory rescue experiment.} The modification removed the modal filter, training on all 2,000 calibration items including those with bimodally distributed self-consistency (84.6\% at $n_\text{correct} = 0$ or 10). This produced a binary verbal correctness discriminator with AUROC$_2$ = 0.774 on held-out TriviaQA. This compresses a 10-sample self-consistency signal (AUROC$_2$ = 0.999) into a single-pass verbal readout that exceeds single-pass logit entropy (AUROC$_2$ = 0.701), with a compression loss of approximately 22.5\%. The ceiling rate collapsed from 97.7\% to 49.8\%. VRS improved from Invalid to Indeterminate. The shuffled-target control showed no improvement (AUROC$_2$ = 0.501), confirming the effect is not format learning.

On MMLU, a benchmark absent from training, the real-target model improved accuracy from 54.2\% to 77.4\% and AUROC$_2$ from 0.535 to 0.616. The shuffled-target model remained at baseline (accuracy 56.1\%, AUROC$_2$ = 0.523, parse rate 66/498 vs 470/498 for the real-target model), supporting a target-dependent interpretation of the MMLU improvement rather than a generic LoRA effect, though the low shuffled parse rate (66/498) prevents a fully controlled comparison.

The post-hoc result is exploratory and requires replication. The model produces binary confidence (5\% or 95\%) rather than continuous calibration, and TriviaQA accuracy drops 7.5 percentage points. The result demonstrates that CSFT can distil multi-sample self-consistency into a single-pass verbal signal that discriminates held-out correctness, and identifies two design lessons: confidence training requires examples of low confidence (which the modal filter eliminated), and correct confidence targets regularise output format as a side effect.

\noindent\textbf{Pre-registration:} Open Science Framework (\url{https://osf.io/mpcr5}, filed prior to baseline characterisation).\\
\textbf{Code and data:} \url{https://github.com/synthiumjp/metacog-engineering}
\end{abstract}

\section{Introduction}

\subsection{The verbal confidence problem}

Large language models produce verbal confidence estimates that are systematically uninformative. Under minimal elicitation (asking a model to state its confidence as a percentage), instruct-tuned models at the 3--9B scale exhibit ceiling rates above 90\%, meaning they report near-maximum confidence regardless of whether they are correct \citep{cacioli2026g}. The CMM programme has documented this across seven frontier models using Type-2 signal detection theory (SDT; \citealp{fleming2014}; \citealp{maniscalco2012}). In every case, the verbal confidence signal was classified as Invalid under the Validity Rating Scale (VRS).

This finding is consistent with broader work on LLM confidence. \citet{tian2023} showed that verbalized confidence from RLHF models can be better calibrated than conditional probabilities, but remains systematically overconfident under naive elicitation. \citet{xiong2024} systematically evaluated whether LLMs can express their uncertainty and found persistent miscalibration across elicitation methods.

The degenerate verbal channel coexists with informative internal representations. Linear probes trained on hidden states achieve AUROC$_2$ values in the 0.6--0.8 range on the same items where verbal confidence is near chance \citep{cacioli2026g}. Single-pass logit entropy also discriminates correctness substantially better than verbal confidence. The information exists. The verbal channel does not transmit it.

Recent mechanistic work has begun to characterise why the verbal channel fails. \citet{kumaran2026} showed that verbal confidence is computed automatically during answer generation, cached at answer-adjacent positions, and retrieved for output, with cached representations explaining variance in verbal confidence beyond token log-probabilities. \citet{miao2026} found that calibration and verbalized confidence signals are encoded linearly but are geometrically orthogonal in the residual stream, a dissociation they term the confidence-faithfulness gap. \citet{zhao2026} identified specific circuits that causally inflate verbalized confidence on incorrect answers. Together, these findings suggest that the degenerate verbal channel is not a capacity limitation but a readout failure: the internal signal exists but the generation pathway does not transmit it faithfully.

\subsection{Why the gap matters}

The gap between internal information and verbal readout has practical consequences for any deployment where a model's self-assessment is used for downstream decisions: selective prediction, deferral to human experts, risk flagging, or autonomous agent behaviour. If the verbal channel is degenerate, these applications must rely on logit-level signals (entropy, softmax margins) or external sampling methods (self-consistency), both of which require either model internals access or multiple forward passes. A reliable verbal readout would be cheaper, simpler to deploy, and accessible through any API.

\subsection{Training objectives and confidence quality}

The CMM programme has established that training objectives are the primary determinant of metacognitive readout quality. The bPC paper \citep{cacioli2026l} showed that cross-entropy at the output is empirically load-bearing for the relationship between an energy-based structural probe and softmax-derived confidence. The quantisation paper \citep{cacioli2026e} showed that supervised fine-tuning reshapes confidence distributions without improving metacognitive sensitivity (meta-$d'$). The Atlas paper \citep{cacioli2026c} showed that metacognitive monitoring quality is dissociable from accuracy and scale.

Recent work has begun to address confidence quality through targeted interventions. \citet{wang2026} use self-generated distractors to calibrate verbalized confidence. \citet{li2025} propose ConfTuner, a Brier-score-style fine-tuning objective for confidence. \citet{taubenfeld2025} demonstrate that confidence and self-consistency are related but dissociable signals. \citet{seo2026} identify answer-independence as a primary driver of overconfidence and introduce a fine-tuning framework (ADVICE) that promotes answer-grounded confidence estimation. These approaches share the intuition that the verbal confidence channel is trainable. They differ in the target signal and the evaluation methodology.

\subsection{This study}

We test whether confidence-conditioned supervised fine-tuning (CSFT) produces a verbal confidence signal that discriminates item-level correctness on held-out data. The confidence targets are derived from 10-sample self-consistency at $T = 0.7$: the number of samples that produce a correct answer is mapped to a confidence percentage. We train with LoRA on Gemma~3 4B-it \citep{gemma2025} and evaluate with the CMM programme's psychometric pipeline.

The study was pre-registered as a Phase~0 feasibility spike with a five-outcome decision tree committed before any evaluation data were collected. The pre-registered protocol produced a negative result. A post-hoc modification, removing a training-set filter, produced a strong positive. We report both results, with the post-hoc status clearly marked throughout.

\paragraph{Scope boundary.} The training targets are derived from the model's own sampling distribution, not from an external correctness oracle. A positive result demonstrates that CSFT can produce a verbal signal that discriminates correctness on held-out items. It does not demonstrate that the model's underlying metacognitive capacity has changed. The 10-sample self-consistency signal is a richer input than single-pass logit entropy, so the comparison between the distilled verbal readout and entropy should be understood as comparing a multi-sample-derived signal (expressed verbally after distillation) against a single-pass signal. We use ``metacognitive'' operationally to denote Type-2 discrimination of a readout over its own Type-1 correctness, not to imply human-like introspective access.

\section{Method}

\subsection{Model and hardware}

All experiments used Gemma~3 4B-it \citep{gemma2025} in bfloat16 precision on an AMD Radeon RX 7900 GRE (16GB VRAM) with ROCm PyTorch 2.8.0. Hardware-specific adaptations included eager attention (SDPA kernels not compiled for gfx1100) and direct GPU placement (accelerate's device-map hooks produced asynchronous HIP errors). These adaptations do not affect the analysis pipeline or the decision rules.

\subsection{Data partitioning}

All items were drawn from TriviaQA rc.nocontext validation (17,944 items; \citealp{joshi2017}) and MMLU \citep{hendrycks2021}. A programmatic disjointness filter excluded 524 items used in a prior saturation study \citep{cacioli2026g}. The remaining 17,420 items were shuffled with a fixed seed and sliced contiguously:

\begin{itemize}
\item \textbf{T-eval:} 1,000 items (held-out evaluation, TriviaQA)
\item \textbf{T-cal:} 2,000 items (calibration and training, TriviaQA)
\item \textbf{Step~0:} 500 items (substrate pre-check, TriviaQA)
\item \textbf{M-eval:} 498 MMLU items stratified across six domains
\end{itemize}

Pairwise disjointness was verified programmatically.

\subsection{Baseline characterisation (Step~1)}

Greedy generation with the base model on T-eval and M-eval. Each item was prompted with a minimal elicitation format asking for an answer followed by a confidence percentage. AUROC$_2$ was computed as the area under the receiver operating characteristic curve for confidence as a predictor of correctness. VRS screening classified the confidence signal as Valid, Indeterminate, or Invalid. First-token entropy was computed from the output logits as a single-pass baseline (E5).

A linear probe (L2-regularised logistic regression, 5-fold cross-validation) was fit on hidden states from the T-cal greedy pass and evaluated on T-eval across a $3 \times 2$ grid of layer positions (first/middle/last) $\times$ token positions (pre-answer/last-answer).

\subsection{Calibration and target derivation (Step~2)}

For each T-cal item, 10 responses were generated at $T = 0.7$. The number of correct responses ($n_\text{correct}$) was mapped to a confidence target: $\{0 \to 5\%, 1 \to 15\%, 2 \to 25\%, \ldots, 9 \to 90\%, 10 \to 95\%\}$. Difficulty bins were assigned as Easy ($n_\text{correct} \geq 8$), Medium ($4 \leq n_\text{correct} \leq 7$), and Hard ($n_\text{correct} \leq 3$).

The pre-registered protocol applied a \textbf{modal filter}: only items where the most frequent answer across 10 samples was correct ($\text{modal\_correct} = \text{True}$) were included in the training set. The rationale was to ensure training examples contained correct answers, avoiding potential degradation from training on incorrect content.

\subsection{Fine-tuning (Step~3)}

LoRA fine-tuning (rank 16, $\alpha = 32$, dropout 0.05) on seven target modules (q, k, v, o, gate, up, down projections). Learning rate $2 \times 10^{-4}$ with cosine schedule, 3 epochs, effective batch size 16 via gradient accumulation. The training input was the chat-formatted prompt with the modal answer and confidence target in the assistant turn. Labels were masked on prompt tokens. LoRA hyperparameters were fixed in the pre-registration without optimisation.

A shuffled-target control (seed 43) was trained with identical architecture and hyperparameters but with confidence targets randomly permuted across items. The real-shuffled target correlation was verified below $|r| < 0.05$ before training.

\subsection{Post-hoc modification: removing the modal filter}

The pre-registered protocol produced a negative result (\S\ref{sec:prereg_result}). Post-hoc analysis identified a label-entropy collapse (\S\ref{sec:label_entropy}) as the cause. The modification: all 2,000 T-cal items were included in the training set, including 893 items with incorrect modal answers. The same hyperparameters, shuffled control, and evaluation pipeline were applied. This modification was not pre-registered.

\subsection{Evaluation (Step~4)}

The fine-tuned model (real-target) and shuffled-target control were evaluated on T-eval and M-eval with greedy generation. AUROC$_2$, VRS, accuracy, and ceiling rate were computed. Paired-bootstrap CIs (10,000 resamples) were computed on the AUROC$_2$ delta. The shuffled-target control was evaluated on both TriviaQA and MMLU.

\paragraph{MMLU parsing.} The base model produces MMLU answers in the format ``Answer: C'' (uppercase). The post-SFT model produces ``c. answer text'' (lowercase with content). The shuffled model produces mostly long-form explanations with low parse rate (66/498). Each model was parsed with a format-appropriate parser. The base model's parser was verified with zero mismatches across all 498 responses. The base model never produced lowercase-initial responses (0/498). These parsing details and the format divergence across conditions are reported transparently as they bear on the interpretation of cross-benchmark accuracy (\S\ref{sec:mmlu}).

\subsection{Distinction: discrimination vs.\ calibration vs.\ metacognitive sensitivity}

\textbf{Discrimination} (AUROC$_2$) measures whether a confidence signal rank-orders correct and incorrect responses. \textbf{Calibration} measures whether stated confidence matches empirical accuracy. \textbf{Metacognitive sensitivity} (meta-$d'$, M-ratio) measures the efficiency of confidence as a monitor of correctness relative to Type-1 information.

This study addresses discrimination. The post-hoc result produces binary confidence (5\% or 95\%), which constitutes a two-bin classifier rather than a calibrated probability estimator. The term ``calibration'' is avoided except where accuracy-by-bin data are specifically reported. Meta-$d'$ computation is reserved for the planned scale-up study where continuous confidence output is expected.

\section{Results}

\subsection{Baseline}

The base model exhibited the expected degenerate confidence profile. On T-eval: AUROC$_2$ = 0.554 (near chance), ceiling rate = 97.7\%, VRS = Invalid, accuracy = 57.2\%. On M-eval: AUROC$_2$ = 0.535, VRS = Invalid, accuracy = 54.2\%. Single-pass logit entropy achieved AUROC$_2$ = 0.701.

The probe grid showed AUROC$_2$ ranging from 0.619 (first layer, last-answer token) to 0.837 (middle layer, last-answer token). The primary probe configuration (last layer, last-answer token) achieved 0.805. All six configurations exceeded verbal AUROC$_2$ with CIs excluding zero.

\subsection{Self-consistency distribution and the label-entropy collapse}
\label{sec:label_entropy}

\begin{figure}[t]
\centering
\includegraphics[width=0.85\linewidth]{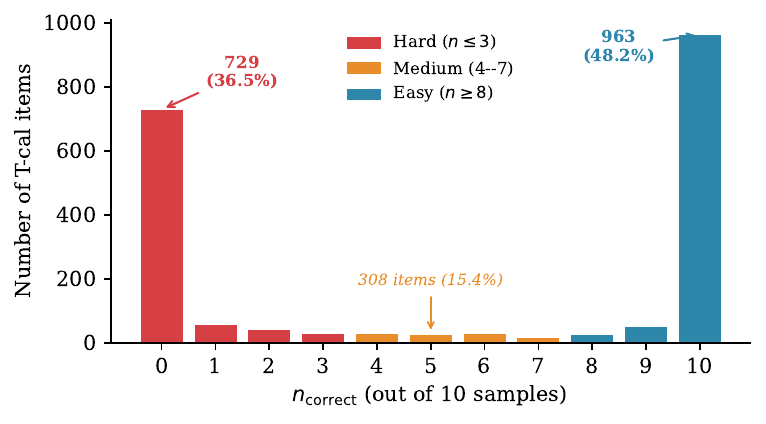}
\caption{Self-consistency distribution on T-cal (2,000 items). The strongly bimodal structure at 4B scale concentrates 84.6\% of items at the extremes ($n_\text{correct} = 0$ or 10), with only 15.4\% in the intermediate range.}
\label{fig:sc_dist}
\end{figure}

The T-cal sampling revealed a strongly bimodal distribution: 963 items (48.2\%) at $n_\text{correct} = 10$ and 729 items (36.5\%) at $n_\text{correct} = 0$. Only 308 items (15.4\%) fell in the intermediate range ($n_\text{correct}$ 1--9). Figure~\ref{fig:sc_dist} shows the full distribution.

Raw 10-sample self-consistency, evaluated as a retrospective signal on T-eval, achieved AUROC$_2$ = 0.999. The near-perfect discrimination reflects the bimodal structure: items the model always gets right ($n_\text{correct} = 10$) are almost always correct on any given attempt, and items it always gets wrong ($n_\text{correct} = 0$) are almost always incorrect. Self-consistency at this model scale is a near-binary oracle.

\begin{figure}[t]
\centering
\includegraphics[width=\linewidth]{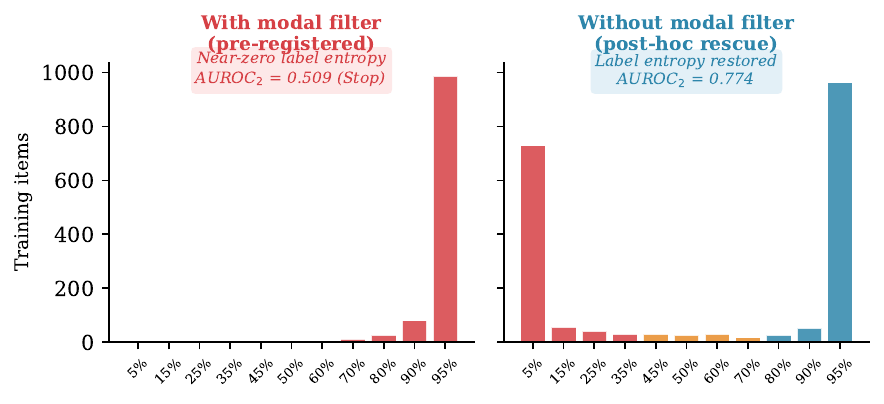}
\caption{Label-entropy collapse. Left: with the modal filter, training targets are near-uniformly high confidence. Right: removing the filter restores label entropy, providing both high- and low-confidence training signal.}
\label{fig:label_entropy}
\end{figure}

Under the modal filter, 1,107 items entered the training set (all $\text{modal\_correct} = \text{True}$, predominantly $n_\text{correct} = 10$ with target 95\%), while 893 items were excluded (predominantly $n_\text{correct} = 0$ with target 5\%). The resulting training target distribution had near-zero entropy: the model was trained almost exclusively on high-confidence targets. This is a label-entropy collapse (Figure~\ref{fig:label_entropy}). The label distribution provides no information about when to output low confidence, so the model's optimal strategy is to output the mode (95\%) regardless of input. The failure of the pre-registered protocol (\S\ref{sec:prereg_result}) follows directly from this distributional property.

\subsection{Pre-registered result (with modal filter): Stop}
\label{sec:prereg_result}

\begin{table}[h]
\centering
\caption{Pre-registered protocol results (with modal filter).}
\label{tab:prereg}
\begin{tabular}{lccc}
\toprule
Metric & Baseline & Post-SFT & Shuffled \\
\midrule
AUROC$_2$ (T-eval) & 0.554 & 0.509 & 0.501 \\
Accuracy (T-eval) & 0.572 & 0.496 & --- \\
Ceiling rate & 0.977 & 0.984 & --- \\
VRS & Invalid & Invalid & --- \\
\bottomrule
\end{tabular}
\end{table}

H1: $\delta = -0.052$, 95\% CI $[-0.077, -0.027]$ (Table~\ref{tab:prereg}). \textbf{Not met.} Decision tree terminal state: \textbf{Stop.}

\subsection{Post-hoc result (no modal filter): exploratory}

All results in this section are from the exploratory post-hoc modification and were not pre-registered.

\subsubsection{TriviaQA (T-eval)}

\begin{table}[h]
\centering
\caption{Post-hoc results on TriviaQA (no modal filter).}
\label{tab:posthoc}
\begin{tabular}{lccc}
\toprule
Metric & Baseline & Post-SFT (no filter) & Shuffled \\
\midrule
AUROC$_2$ & 0.554 & \textbf{0.774} & 0.501 \\
Accuracy & 0.572 & 0.497 & --- \\
Ceiling rate & 0.977 & \textbf{0.498} & --- \\
VRS & Invalid & \textbf{Indeterminate} & --- \\
AURC & 0.377 & \textbf{0.309} & --- \\
\bottomrule
\end{tabular}
\end{table}

H1: $\delta = 0.168$, 95\% CI $[0.132, 0.203]$ (Table~\ref{tab:posthoc}). The shuffled control did not move (0.501), confirming the effect is not format learning.

The confidence output was effectively binary: 494 items at 5\% and 498 at 95\%. Items assigned 95\% confidence had 77.1\% accuracy. Items assigned 5\% confidence had 22.3\% accuracy. The learned signal is equivalent to a two-bin correctness classifier rather than a probabilistic estimator.

\paragraph{Distillation framing.} The intervention compresses a 10-sample self-consistency signal (AUROC$_2$ = 0.999) into a single-pass verbal readout (AUROC$_2$ = 0.774), a compression loss of approximately 22.5\%. The distilled readout exceeds single-pass logit entropy (AUROC$_2$ = 0.701) by 0.073 points (Table~\ref{tab:signals}). The practical implication is that multi-sample uncertainty information can be distilled into a verbal signal accessible in a single greedy decode, at a cost of approximately one-quarter of the available discrimination.

\begin{table}[h]
\centering
\caption{Signal comparison by inference cost.}
\label{tab:signals}
\begin{tabular}{lcc}
\toprule
Signal & AUROC$_2$ & Inference cost \\
\midrule
Self-consistency (10 samples) & 0.999 & 10 forward passes \\
Distilled verbal (CSFT) & 0.774 & 1 forward pass \\
Logit entropy & 0.701 & 1 forward pass \\
Baseline verbal & 0.554 & 1 forward pass \\
Shuffled verbal & 0.501 & 1 forward pass \\
\bottomrule
\end{tabular}
\end{table}

TriviaQA accuracy dropped 7.5 percentage points (57.2\% $\to$ 49.7\%). This drop means the intervention changed answer behaviour, not just confidence reporting (see \S\ref{sec:confound}).

\begin{figure}[t]
\centering
\includegraphics[width=0.85\linewidth]{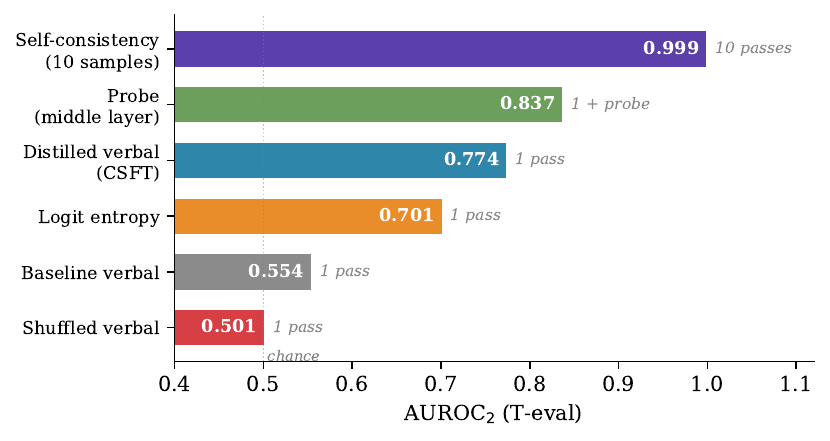}
\caption{Correctness discrimination (AUROC$_2$) by signal type and inference cost. The distilled verbal readout (CSFT) recovers approximately 77.5\% of the 10-sample self-consistency signal in a single forward pass.}
\label{fig:signals}
\end{figure}

\subsubsection{Within-bin analysis (H3)}

H3 was not met. The Easy bin produced degenerate bootstrap (uniform confidence). The Hard bin showed a positive trend ($\delta = 0.176$, CI $[-0.214, 0.374]$). The Medium bin moved in the negative direction ($\delta = -0.233$). The binary confidence output provides between-category separation but no within-bin rank ordering.

\subsubsection{MMLU (M-eval): cross-benchmark transfer with shuffled control}
\label{sec:mmlu}

\begin{table}[h]
\centering
\caption{MMLU cross-benchmark results.}
\label{tab:mmlu}
\begin{tabular}{lccc}
\toprule
Condition & Accuracy & AUROC$_2$ & Parse rate \\
\midrule
Baseline & 54.2\% & 0.535 & 498/498 \\
Real-target CSFT & \textbf{77.4\%} & \textbf{0.616} & 470/498 \\
Shuffled-target & 56.1\% & 0.523 & 66/498 \\
\bottomrule
\end{tabular}
\end{table}

The real-target model improved MMLU accuracy by 23.2 percentage points and AUROC$_2$ by 0.081 compared to baseline (Table~\ref{tab:mmlu}). The shuffled-target model remained at baseline accuracy (56.1\%) and AUROC$_2$ (0.523), with a substantially lower parse rate (66/498 vs 470/498). This dissociation confirms that the MMLU improvement is attributable to the confidence target information, not the LoRA fine-tuning procedure itself.

The three conditions produced markedly different response formats. The baseline produces structured ``Answer: X'' responses (498/498 parseable). The real-target model produces concise ``x.\ answer text\textbackslash nConfidence: N\%'' responses (470/498 parseable). The shuffled-target model produces verbose explanations with occasional embedded answers (66/498 parseable). Parser verification confirmed zero mismatches on the baseline parser, and the baseline never produced lowercase-initial responses (0/498 items). The format divergence between real-target and shuffled conditions, both trained on identical answer content, suggests that correct confidence targets may regularise output format as a side effect.

\paragraph{Interpretation.} The MMLU accuracy improvement is suggestive and requires replication. The shuffled-target control establishes that the improvement is target-driven, not a LoRA artifact. Three interpretive hypotheses remain: (1) correct confidence targets reduce overconfident confabulation, improving answer selection on constrained tasks; (2) the confidence-answer co-training produces a response format that better elicits the model's knowledge; (3) correct targets regularise output structure in ways that incidentally benefit multiple-choice performance. We report the result transparently without claiming domain-general epistemic improvement.

\subsection{Training dynamics}

Both real-target and shuffled-target runs converged to similar final loss ($\sim$0.11 after 3 epochs on 2,000 items). The model memorised both target mappings equally well. The critical difference was generalisation: the real-target model's confidence discriminates correctness on held-out items and produces concise, parseable responses. The shuffled-target model's confidence is non-discriminative and its responses are verbose and poorly structured.

\section{Discussion}

\subsection{Label-entropy collapse: why the modal filter guarantees failure}

The modal filter's failure can be understood as a label-entropy problem. The training target distribution under the filter had near-zero entropy: of 1,107 training items, the overwhelming majority had target 95\%. The model's loss-minimising strategy under a near-constant target distribution is to ignore the input and output the modal target value. This produces the observed behaviour: post-SFT confidence is uniformly high, indistinguishable from the baseline ceiling effect.

The no-filter design restores label entropy. With 729 items at target 5\% and 963 at target 95\%, the model must learn to discriminate inputs to minimise loss. The bimodal distribution concentrates training at the two extremes, which is why the model learns a binary discriminator rather than a continuous mapping. The binary discriminator is the optimal solution given the available training signal.

This analysis yields a concrete design principle: \textbf{confidence training requires label entropy.} Any training-set filter that removes low-confidence examples will collapse the label distribution and guarantee failure. When the goal is to train a model to report uncertainty, the examples of uncertainty are the essential training signal.

\subsection{Self-consistency distillation}

The intervention compresses a 10-sample self-consistency signal into a single-pass verbal readout. Raw self-consistency achieves AUROC$_2$ = 0.999 on T-eval. This is near-perfect but requires 10 forward passes at inference time. The distilled verbal readout achieves 0.774 in a single pass, recovering approximately 77.5\% of the available discrimination at one-tenth the inference cost.

The distilled readout exceeds single-pass logit entropy (0.701). This does not indicate that the verbal channel has acquired information beyond what the model internally possesses. The training targets encoded 10-sample information that entropy, by construction, does not capture. The contribution is practical. Multi-sample uncertainty information, once distilled, is accessible through a standard API call without logit access or repeated sampling.

\subsection{The accuracy-policy confound}
\label{sec:confound}

TriviaQA accuracy dropped 7.5 percentage points (57.2\% $\to$ 49.7\%). This study does not establish that monitoring improved independently of policy. It establishes that a joint answer-confidence policy can be trained to produce a discriminative verbal signal. The improved AUROC$_2$ may reflect better monitoring of a changed answer distribution, or a policy that simultaneously produces easier-to-classify answer-confidence pairs.

The shuffled-target control partially addresses this confound: the shuffled model underwent the same LoRA procedure and saw the same answer content. Its confidence is non-discriminative (AUROC$_2$ = 0.501). If the accuracy drop alone drove the discrimination, the shuffled model should show similar discrimination with similar accuracy loss. It does not. A definitive resolution would require a design that isolates confidence from answer generation, for example training the model to reproduce the base model's greedy answer verbatim and append a calibrated confidence value.

\subsection{MMLU accuracy improvement: three hypotheses}

The 23.2 percentage point MMLU accuracy improvement (54.2\% $\to$ 77.4\%) is attributable to the confidence targets (shuffled control stays at baseline) but the mechanism remains unclear.

\paragraph{Hypothesis 1: Reduced confabulation.} The base model's overconfidence may cause it to commit prematurely to wrong answers. The CSFT training, which explicitly associates low confidence with incorrect responding, may make the model more cautious and more likely to select the correct option on constrained multiple-choice tasks.

\paragraph{Hypothesis 2: Format regularisation.} Correct confidence targets produced concise ``letter.\ content\textbackslash nConfidence: N\%'' responses (470/498 parseable). Shuffled targets produced verbose explanations (66/498 parseable). The concise format may better elicit the model's existing knowledge by reducing the opportunity for reasoning errors in long-form generation.

\paragraph{Hypothesis 3: Confidence-answer co-training.} Training on correct confidence targets may create a joint policy where high confidence and correct answers are mutually reinforcing. The model may learn that ``95\% confidence'' co-occurs with correct answers and adjusts its answer selection accordingly.

These hypotheses are not mutually exclusive. Resolving them requires ablations beyond the scope of this Phase~0 study: format-controlled evaluations, a second MCQA benchmark, and analysis of the LoRA's effect on intermediate representations.

\subsection{Limitations}

Several limitations constrain interpretation. The confidence output is binary (two-bin classifier, not continuous calibration). TriviaQA accuracy drops despite improved discrimination. The study used a single model, single seed, and single benchmark pair. The within-bin analysis (H3) failed due to binary confidence providing no within-bin rank ordering. The positive result is post-hoc. The MMLU improvement, though confirmed by shuffled control, involves a format change whose full implications are not characterised. Meta-$d'$ and continuous calibration metrics (Brier score, ECE) are not computed and are reserved for the scale-up study.

\section{Threats to validity}

\textbf{Threat 1: Post-hoc rescue.} The positive result is exploratory. Replication with pre-registered no-filter design is needed.

\textbf{Threat 2: Single model, single seed.} Results may not generalise across model families or random initialisations.

\textbf{Threat 3: Binary confidence output.} AUROC$_2$ is maximised by two well-separated bins. The high value partly reflects this structure.

\textbf{Threat 4: Accuracy-policy confound.} The intervention changed answer behaviour, not just confidence reporting. The shuffled control partially but not fully resolves this.

\textbf{Threat 5: Entropy comparison fairness.} The distilled signal derives from 10 samples. The entropy baseline is single-pass. The comparison is multi-sample-derived vs single-pass, not equal-information.

\textbf{Threat 6: MMLU format interaction.} The real-target model's concise format may contribute to accuracy gains independently of confidence quality.

\textbf{Threat 7: Bimodal training distribution.} 84.6\% of training items had extreme $n_\text{correct}$ values. The model may identify surface features of ``always correct'' vs ``always incorrect'' items rather than estimating graded uncertainty.

\textbf{Threat 8: No meta-$d'$ or calibration metrics.} AUROC$_2$ measures discrimination only. Metacognitive sensitivity and continuous calibration are not evaluated.

\section{Conclusion}

A pre-registered CSFT protocol with a modal filter failed due to label-entropy collapse in the training target distribution. An exploratory no-filter variant produced a binary verbal correctness discriminator on held-out TriviaQA (AUROC$_2$ = 0.774), compressing a 10-sample self-consistency signal (AUROC$_2$ = 0.999) into a single-pass readout that exceeds logit entropy (0.701). The shuffled-target control confirmed the effect is target-driven on both TriviaQA (shuffled AUROC$_2$ = 0.501) and MMLU (shuffled accuracy at baseline, parse rate 66/498 vs 470/498).

The result establishes two design principles: first, confidence training requires label entropy. Filtering out low-confidence examples guarantees failure. Second, correct confidence targets appear to regularise output format, producing concise, parseable responses where shuffled targets produce verbose, poorly structured output.

The result is exploratory, binary rather than continuously calibrated, and observed at a single model scale with a single seed. Replication at larger scale, where the self-consistency distribution may smooth out and the accuracy trade-off may resolve, is the necessary next step.

\bibliographystyle{apalike}

\end{document}